\documentclass[conference]{IEEEtran}
\IEEEoverridecommandlockouts

\usepackage{amsmath}
\usepackage{amssymb}
\usepackage{amsthm}
\usepackage{bm}
\usepackage{graphicx}
\usepackage{subcaption}
\usepackage{floatrow}
\usepackage{xspace}
\usepackage{blindtext}
\usepackage{hyperref}
\usepackage{algpseudocode}
\usepackage{algorithm}
\usepackage{mathtools}
\usepackage{times}
\usepackage{tikz}
\usepackage{pgfplots}
\usepackage{tabularx}
\usepackage{booktabs}
\pgfplotsset{compat=1.18}

\DeclareMathOperator*{\argmin}{arg\,min}

\long\def\comment#1{}
\long\def\red#1{\bgroup\color{red}#1\egroup}

\newcommand{\name}{ALPCAH }
\newcommand{\secondname}{ALPCAH}

\newcommand{\xmath}[1] {\ensuremath{#1}\xspace}

\newcommand{\normfrob}[1] {\xmath{\left\| #1 \right\|_{\mathrm{F}}}} 
\newcommand{\normfrobr}[1] {\xmath{\| #1 \|_{\mathrm{F}}}} 

\newcommand{\pbox}[1] {%
\makebox[0pt][r]{\raisebox{7mm}[0pt][0pt]{\small #1}}\ignorespaces}
\newcommand{\be} {\begin{equation}}
\newcommand{\ee}[1] {\label{#1}\end{equation}\pbox{#1}}

\newcommand{\fref}[1] {Fig.~\ref{#1}}

\newcommand{\bs}{\begin{equation}\begin{split}}

\newcommand{\st}{\hspace{2mm} \text{s.t.} \hspace{2mm}}

\renewcommand{\xi}{\xmath{x_i}}

\newtheorem{definition}{Definition}
\newtheorem{theorem}{Theorem}

\ifCLASSINFOpdf
\else
\fi
\begin{document}
%
\title{
\secondname: Sample-wise Heteroscedastic PCA
\\
with Tail Singular Value Regularization

}
%


\author{
 \IEEEauthorblockN{ Javier Salazar Cavazos, Jeffrey A. Fessler, Laura Balzano}%
 \thanks{Supported in part by NSF CAREER Grant 1845076.}
\IEEEauthorblockA{
EECS Department,
University of Michigan, Ann Arbor, Michigan, United States\\
Email: javiersc@umich.edu, fessler@umich.edu, girasole@umich.edu}
}



%


\maketitle

\begin{abstract}
Principal component analysis (PCA) is a key tool in the field of data dimensionality reduction
that is useful for various data science problems.
However, many applications
involve heterogeneous data that varies in quality
due to noise characteristics associated with different sources of the data.
Methods that deal with this mixed dataset are known as heteroscedastic methods.
Current methods like HePPCAT make Gaussian assumptions of the basis coefficients that may not hold in practice.
Other methods such as Weighted PCA (WPCA) assume the noise variances are known,
which may be difficult to know in practice. 
This paper develops a PCA method
that can estimate the sample-wise noise variances
and use this information in the model
to improve the estimate of the subspace basis associated with the low-rank structure of the data.
This is done without distributional assumptions of the low-rank component
and without assuming the noise variances are known.
Simulations
show the effectiveness of accounting for such heteroscedasticity in the data,
the benefits of using such a method with all of the data versus retaining only good data,
and comparisons are made against other PCA methods established in the literature
like PCA, Robust PCA (RPCA), and HePPCAT. Code available at
\url{https://github.com/javiersc1/ALPCAH}.
\end{abstract}



\section{Introduction}
Many modern data science problems require
learning an approximate subspace basis
for some collection of data.
For example,
lesion detection \cite{lesion_detection},
motion estimation \cite{motion_estimation},
dynamic MRI \cite{dynamic_mri},
and image/video denoising \cite{video_denoising}
are practical applications involving the estimation of a subspace basis. 
Today, a voluminous amount of data is collected to solve problems
and this data tends to have a high dimensional ambient space.
However, the underlying relationships between the variables
are often low dimensional
so the problem becomes finding low dimensional structure in the data
to achieve a certain task.

PCA methods like Robust PCA \cite{RPCA} and Probabilistic PCA \cite{ppca}
work well in the homoscedastic setting,
i.e., when the data is the same quality throughout,
but fail to accurately estimate the basis
when the data varies in quality,
i.e., in the heteroscedastic setting \cite{heppcat}.
In this setting, the noisier data samples can wildly corrupt the basis estimate.
Some examples of heteroscedastic data sets
that involve subspace bases
include environmental air data \cite{epa}, astronomical image data \cite{astronomical_data},
and biological sequencing data \cite{bacteria_data}.
A natural question to ask is whether it is possible
to simply discard the noisy samples to avoid this issue.
This question assumes that the practitioner knows what samples are good and bad,
which may be difficult to know in practice.
The question also assumes that there is enough good data to estimate the basis,
but it is possible that the general lack of good data 
requires using the noisy data if the subspace dimension is higher than the amount of good data.
More optimistically,
even the noisier samples can help improve the estimate of the basis
if properly modeled
\cite{heppcat},
so it is preferable to use all of the data available.  

Although one can consider heteroscedasticity across the feature space
with methods such as HeteroPCA \cite{hetero_pca},
this paper focuses on heteroscedasticity across the data samples.
The weighted PCA (WPCA) \cite{wpca}
approach for heteroscedastic data
forms a weighted sample covariance matrix and requires knowledge of the noise variances.
However, data quality may not be known,
e.g., unknown origin of the dataset or unavailable data sheet for physical sensors.
Other heteroscedastic methods like HePPCAT \cite{heppcat} use factor analysis
and hard rank constraints to estimate the subspace basis.
Being a probabilistic PCA approach,
HePPCAT makes Gaussian assumptions about the basis coefficients. 
Additionally, HePPCAT either assumes the subspace dimension is known
or requires an estimate of the rank parameter.
The proposed method in the next section
allows for optional usage of rank knowledge via a unique low-rank promoting functional
and makes no distributional assumptions about the low-rank component,
allowing it to achieve higher accuracy than current methods
even without knowing the noise variances.

\section{Proposed Method}

Let $y_i \in \mathbb{R}^{D}$ represent the data samples
for index $i \in \{1,\ldots,N \}$ given $N$ total samples,
and let $D$ denote the ambient dimension.
Let $x_i$ represent the low-dimensional data sample
generated by $x_i = U z_i$
where $U \in \mathbb{R}^{D \times k}$ is an unknown subspace basis of dimension $k$
and $z_i \in \mathbb{R}^{k}$ are basis coordinates. 
Then the heteroscedastic model is described as follows assuming Gaussian noise
\begin{equation}
y_i = x_i + \epsilon_i \hspace{2mm} \text{where} \hspace{2mm} \epsilon_i \sim \mathcal{N}(0, \nu_i I)
\end{equation}
for noise variances $\nu_i$.
Note that we are considering the general case where each data point has its own noise variance since it is more challenging to tackle. 
However, one can consider groups of data $\{ \nu_1,\ldots,\nu_L \}$ where $L$ represents the number of groups and each data point belongs to one of the $L$ groups.
For the measurement model
$y_i 
\sim \mathcal{N}(x_i, \nu_i I)$,
the probability density function for a single point is
\begin{equation}
 \frac{1}{\sqrt{(2\pi)^k |\nu_i I |}} \exp{ [-\frac{1}{2} (y_i-x_i)^T (\nu_i I)^{-1} (y_i-x_i)] }. 
\end{equation}
For uncorrelated samples, 
the joint log likelihood of all $y_i$ is the following after dropping constants
\begin{equation}
    \sum_{i=1}^{N} -\frac{1}{2} \log |\nu_i I| -\frac{1}{2} (y_i - x_i)^T (\nu_i I)^{-1} (y_i - x_i).
\end{equation}
Let
$\Pi = \mathrm{diag}(\nu_1,\ldots,\nu_N) \in \mathbb{R}^{N \times N}$
be a diagonal matrix representing the (typically unknown) noise variances.
Let
$Y = [y_1, \ldots,y_N] \in \mathbb{R}^{D \times N}$
represent all of the data samples.
Then, the log likelihood in matrix form is
\begin{equation}
    -\frac{D}{2} \log |\Pi| - \frac{1}{2} \text{Trace}[(Y-X)^T \Pi^{-1} (Y-X)].
\end{equation}
Using trace properties,
the optimization problem we pose
for the heteroscedastic model is
\begin{equation}
    \argmin_{X,\Pi} \lambda f_k(X) + \frac{1}{2} \normfrobr{ (Y - X) \Pi^{-1/2} }^2 + \frac{D}{2} \log \underbrace{|\Pi|}_{\hidewidth \text{determinant} \hidewidth}
    \label{eq:cost},
\end{equation}
where $f_k(X)$ is a relatively new functional in the literature \cite{pssv} that promotes low-rank structure in $X$
and
$\lambda \in \mathbb{R}_{+}$ is a regularization parameter.
Our algorithm for solving \eqref{eq:cost}
is called \name
(\textbf{A}lgorithm for \textbf{L}ow-rank regularized \textbf{PCA} for \textbf{H}eteroscedastic data).
Since $X$ represents the denoised data matrix,
the subspace basis is calculated by performing an SVD on the optimal solution from \eqref{eq:cost} so that $\Hat{X} = \sum_i \Hat{\sigma_i} \Hat{u_i} \Hat{v_i}'$ and thus $\Hat{U} = [\Hat{u_1},\ldots,\Hat{u_k}]$.
The low-rank promoting functional we use is the summation of the tail singular values
defined as the following
\begin{equation}
f_k (X) \triangleq \sum_{i=k+1}^{\text{min}(D,N)} \sigma_i (X) = \|X\|_* - \|X\|_{\mathrm{Ky-Fan}(k)}
\end{equation}
where $\sigma_i (X)$ is the ith singular value of $X$,
$\| \cdot \|_*$ is the nuclear norm,
and $\| \cdot \|_{\mathrm{Ky-Fan}(k)}$ is the Ky-Fan norm
defined as the sum of the first $k$ singular values.
For $k=0$, $f_0(X) = \|X\|_*$.
For a general $k>0$, $f_k(X)$ is a nonconvex difference of convex functions.
When $k>0$ and $\lambda \rightarrow \infty$,
then the solution of the optimization problem approaches
$\Hat{X} = \sum_{i=1}^k \sigma_i u_i v_i' \in \mathbb{R}^{D \times k}$
meaning the solution becomes identical to a singular value projection approach. 

\section{Algorithm \& Convergence Analysis}

We apply the inexact augmented Lagrangian method ADMM \cite{admm}
to the cost function \eqref{eq:cost}.
Introducing the auxiliary variable
$Z = Y - X$,
the augmented penalty parameter $\mu \in \mathbb{R}$,
and dual variable $\Lambda \in \mathbb{R}^{D \times N}$,
the augmented Lagrangian, as defined in \cite{alm_theory}, is
\begin{align}
\mathcal{L}_{\mu}(X,Z, & \Lambda, \Pi) =
\lambda_r f_k(X) + \frac{1}{2} \normfrobr{ Z \Pi^{-1/2} } ^2 + \frac{D}{2} \log |\Pi |
\nonumber \\ 
&+ \langle \Lambda, Y-X-Z \rangle + \frac{\mu}{2} \normfrobr{ Y-X-Z } ^2 . \label{eq:auglagrangian}
\end{align}
\begin{definition}
Let $A \in \mathbb{R}^{D \times N}$ be a rank $k$ matrix
such that its decomposition is $\mathrm{SVD}(A) = U_A D_A V_A '$
where $D_A = \mathrm{diag}(\sigma_1 (A), \ldots, \sigma_{\min(D,N)} (A))$.
Let the soft thresholding operation be defined as
$\mathcal{S}_{\tau}[x] = \mathrm{sign}(x) \max(|x| - \tau, 0)$
for some threshold $\tau > 0$.
Decompose $D_A$ such that
$D_A = D_{A1} + D_{A2}
= \mathrm{diag}(\sigma_1 (A),\ldots,\sigma_{k}(A),0,\ldots,0)
+ \mathrm{diag}(0,\ldots,0,\sigma_{k+1}(A),\ldots,\sigma_{N} (A))$.
Then, the proximal mapping solution for $f_k(X)$, as shown in \cite{pssv},
is denoted as the tail singular value thresholding operation and expressed as
\begin{equation}
\mathrm{TSVT}(A, \tau, k) \triangleq U_A \, (D_{A1} + \mathcal{S}_{\tau} [D_{A2}] ) \, V_A'.
\end{equation}
\end{definition}
Performing a block Gauss-Seidel pass for each variable
results in the following closed-form updates
\begin{align}
    Z_{i+1} &= \argmin_{Z_i} \mathcal{L}_{\mu}(X_i,Z_i,\Lambda_i,\Pi_i) \nonumber \\ &= [\mu (Y-X_i) + \Lambda_i](\Pi_i^{-1} + \mu I )^{-1} \label{eq:zupdate}\\
X_{i+1} &= \argmin_{X_{i}} \mathcal{L}_{\mu}(X_i,Z_i,\Lambda_i,\Pi_i) \nonumber \\ &= \mathrm{TSVT}(Y-Z_i+\frac{1}{\mu} \Lambda_i, \frac{\lambda_r}{\mu}, k) \label{eq:xupdate}\\
\Lambda_{i+1} &= \Lambda_i + \mu(Y-X_i-Z_i). \label{eq:lambdaupdate}
\end{align}
When each point is treated as having its own noise variance,
then the variance update is
\begin{align}
    \Pi_{i+1} &= \argmin_{\Pi_i} \mathcal{L}_{\mu}(X_i,Z_i,\Lambda_i, \Pi_i) = \frac{1}{D} Z_i^T Z_i \odot I.
    \label{eq:var_update}
\end{align}
For the case when the data points have grouped noise variances, let
$l \in \{1,\ldots, L\}$ and let $n_l$ signify the number of points in group $l$ out of $L$ total groups; then
the grouped noise variance update instead becomes
\begin{equation}
\nu_l = \frac{1}{D n_l} \normfrobr{ Z^{(p_l)} } ^2 = \frac{1}{D n_l} \normfrobr{ Y^{(p_l)} - X^{(p_l)} } ^2
\end{equation}
where $p_l$ signifies the points associated with group $l$,
meaning that $ Y^{(p_l)} \subset Y $.

Consider the cost function for the case when the variances are known. The formulation consists of a two-block setup written as 
\begin{equation}
\argmin_{X,Z} \underbrace{\lambda_r f_k ( X )}_{f(X)} + \underbrace{\frac{1}{2} \normfrobr{ Z \Pi^{-1/2} } ^2}_{g(Z)}  \st Y = X + Z .
\end{equation}
\begin{theorem}
Let $\Psi(X,Z) = f(X) + g(Z)$. Let $\nu_i \geq \epsilon > 0 \hspace{3mm} \forall i$. Assuming that $\mu$ in \eqref{eq:auglagrangian} satisfies $\mu > 2 L_g = 2 \|\Pi^{-1}\|_2 $, the sequence generated by \eqref{eq:zupdate}, \eqref{eq:xupdate}, \eqref{eq:lambdaupdate} converges to a KKT (Karush–Kuhn–Tucker) point of the augmented Lagrangian $\mathcal{L}_{\mu}(X,Z,\Lambda)$.
\end{theorem}
\begin{proof}
ADMM convergence for nonconvex problems has been studied by \cite{admm_nonconvex} for two-block setups. The functional $f(X)$ is a proper, lower semi-continuous function since it is a sum of continuous functions. The function $g(Z)$ is a continuous differentiable function whose gradient is Lipschitz continuous with modulus of continuity $L_g = \|\Pi^{-1}\|_2$ . Since $g(Z) = \nu_1^{-1/2} Z_{1,1} + \nu_1^{-1/2} Z_{2,1} + \ldots $ is a polynomial equation, then its graph is a semi-algebraic set. 

To the best of our knowledge, there is no literature that explores semi-algebraic properties of nuclear norm based functions and so the following results are our own. Let $f_k(X) = h(X) - q(X) = \|X\|_* - \|X\|_{\mathrm{Ky-Fan}(k)}$. Let $X \in \mathbb{R}^{M \times N}$ such that $G = X'X \in \mathbb{R}^{N \times N}$. Then, by Cayley Hamilton theorem, the characteristic polynomial is expressed as $p_{G} (\lambda) = \lambda^n + c_{n-1}(G) \lambda^{n-1} + \ldots + c_1 (G) \lambda + c_0$ for constants $c_i \in \mathbb{R}$. Let $\lambda$ be eigenvalues of $G$ which implies $p_G (\lambda) = 0$. Then, the set $\mathcal{S}_G = \{ \forall \lambda \hspace{1mm} | \hspace{1mm} p_G (\lambda) = 0 \}$ is semi-algebraic since it is defined by polynomial equations. Note that $\lambda_i = \sigma_i^2$ since $G$ is the gram matrix of $X$. The set $\mathcal{S}_X = \{ \forall \sigma \hspace{1mm} | \hspace{1mm} \sigma^2 = \lambda \in S_G, \sigma \geq 0 \} = \{ (\sigma_1, \ldots, \sigma_n) \}$ is semi-algebraic as it is expressed in terms of polynomial inequalities. Expressing $h(X) = \|X\|_* = h(\sigma_1, \ldots, \sigma_n) $, its graph $h = \{ (\mathbf{\sigma}, f(\mathbf{\sigma})) \} $ is semi-algebraic and thus so is the nuclear norm. By Tarksi-Seidenburg theorem \cite{tarski-seidenburg}, defining the projection map $\Phi : \mathbb{R}^n \rightarrow \mathbb{R}^k$, the set $\Phi (\mathcal{S}_X) = \{ (\sigma_1, \ldots, \sigma_k) \}$ is semi-algebraic and thus so is $q(X) = \|X\|_{\mathrm{Ky-Fan}(k)}$. A finite weighted sum of semi-algebraic functions is known to be semi-algebraic \cite{attouch_presentation} and so $f(X) = h(X)-q(X)$ is semi-algebraic. 

Since the functions $f(X)$ and $g(Z)$ are lower, semi-continuous and definable on an o-minimal structure (such as semi-algebraic or sub-analytic as an example) \cite{o-minimal} then it follows that $\Psi(X,Z) = f(X) + g(Z)$ is a Kurdyka-Łojasiewicz function \cite{attouch_presentation} which is sufficient to proving a bounded sequence. Then the sequence $\{ (X_i,Z_i) \}_{i \in \mathbb{N}}$ converges to a KKT point by applying Theorem 3.1 from \cite{admm_nonconvex}. The unknown variance case involves a challenging nonconvex three-block non-separable setup because of the $g(Z,\Pi)$ term that, to the best of our knowledge, has not been explored in the ADMM literature and thus is a topic of future work.
\end{proof}

\section{Results}
This section uses a synthetic dataset
to compare \name with PCA, RPCA, and HePPCAT.
We consider two groups of data, one with fixed quality (i.e., fixed size and fixed additive noise variance) and one whose parameters we vary.
Let $ y_i \in \mathbb{R}^{100 \times N} $ where $N$, the total number of points,
changes depending on parameter values.
Let $U \in \mathbb{R}^{100 \times 10}$ represent a 10 dimensional subspace
generated by random uniform matrices such that $U \Sigma V^T = \mathrm{svd}(A)$,
where $A_{i,j} \sim \mathcal{U}[0,1]$.
The low-rank data $x_i$ we simulated as $x_i = U z_i$
where the coordinates $z_i \in \mathbb{R}^{10}$
were generated from
$\mathcal{U}[-100,100]$ for each element in the vector.
Then, we generated
$y_i = U z_i + \epsilon_i$
where $\epsilon_i \in \mathbb{R}^{100}$ is drawn from $\mathcal{N}(0, \nu_i I)$.
The noise variance for group 1 ($\nu_1$)
was fixed to 1
and we varied group 2 noise variances ($\nu_2$).
The error metric used is subspace affinity error
that compares the difference in projection matrices
$ \normfrobr{ UU' - \Hat{U} \Hat{U}' } / \normfrobr{ U U' } $
so that a low error signifies a closer estimate of the true subspace. In summary, the noisy data $Y = [y_1,\ldots,y_N]$ is generated accordingly, a solution $\Hat{X}$ is generated from \eqref{eq:cost}, the subspace basis is calculated by $\Hat{X} = \sum_i \Hat{\sigma_i} \Hat{u_i} \Hat{v_i}' \implies \Hat{U} = [\Hat{u_1},\ldots,\Hat{u_k}]$, and the subspace affinity error is reported.

\begin{figure}[H]
         \centering
         \includegraphics[width=1\textwidth]{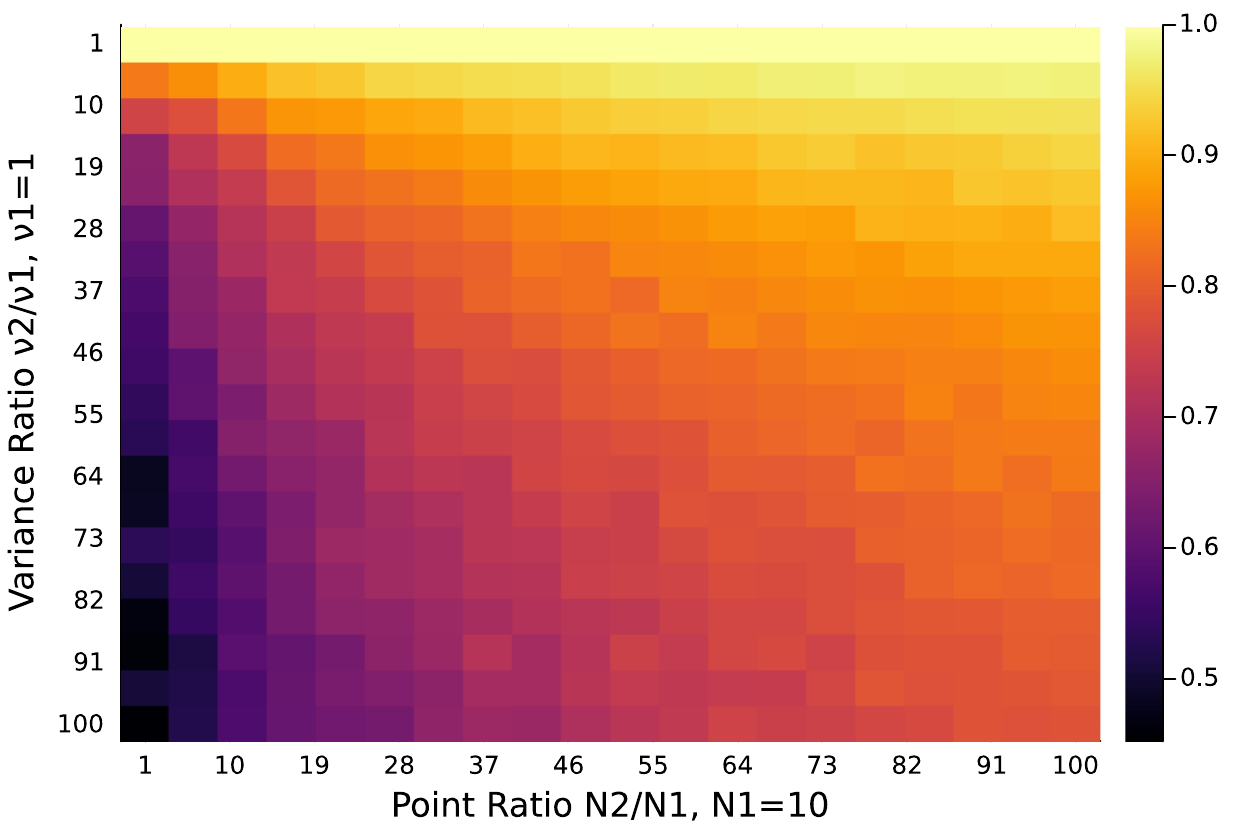}
         \caption{Ratio of subspace affinity errors \secondname/PCA (known variance, no cross-validation required)}
         \label{fig:pca}
\end{figure}

\begin{figure}[H]
         \centering
         \includegraphics[width=1\textwidth]{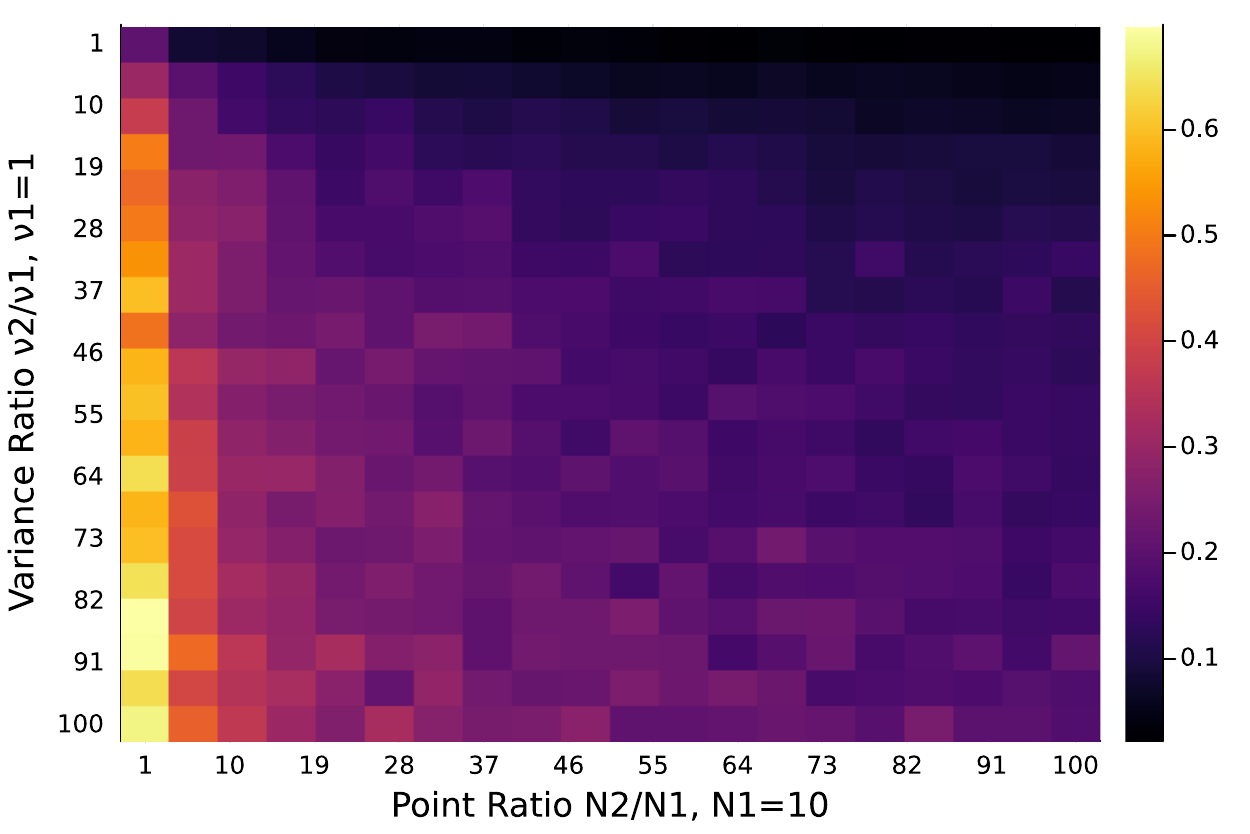}
         \caption{Ratio of subspace affinity errors \secondname/PCA-GOOD (PCA using good data only and \name using all of the data, no cross-validation required)}
         \label{fig:pca_good}
\end{figure}

\begin{figure}[H]
         \centering
         \includegraphics[width=\textwidth]{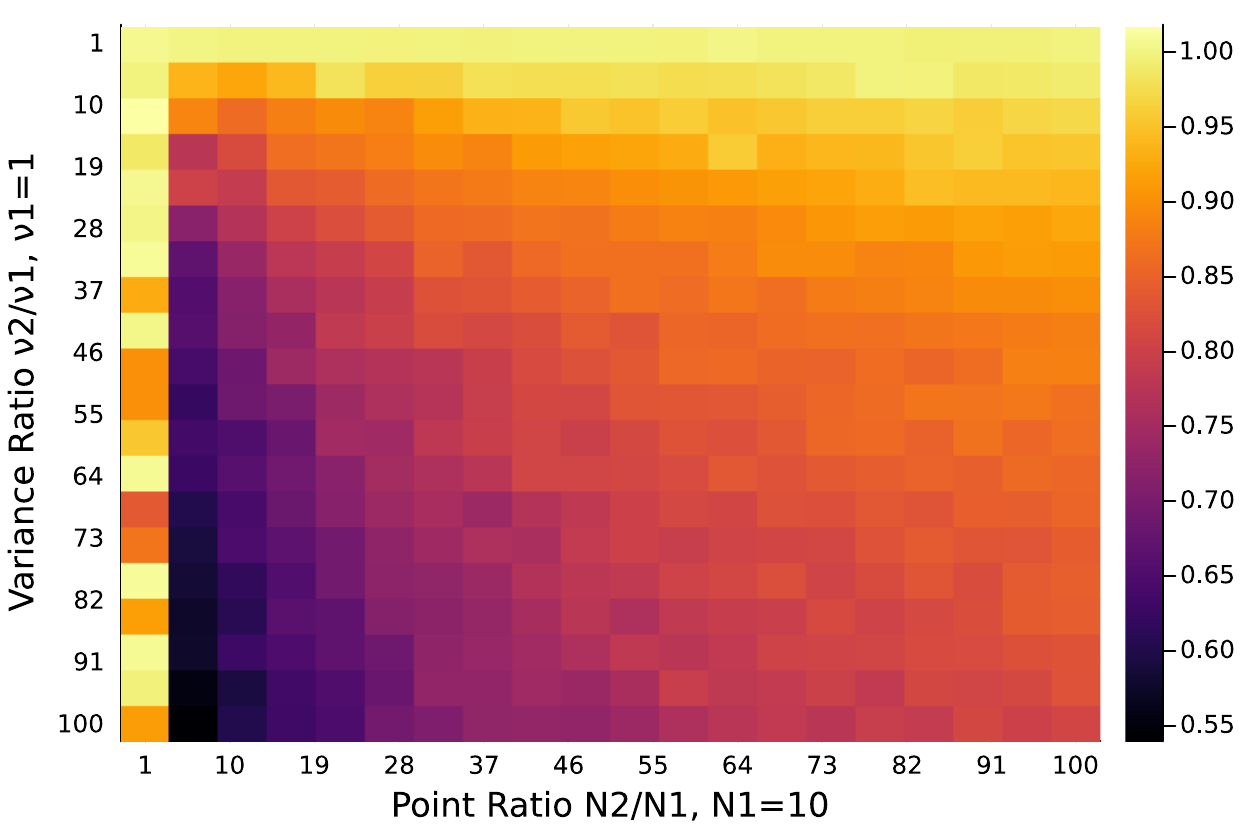}
         \caption{Ratio of subspace affinity errors \secondname/RPCA (unknown variance, no group knowledge, cross-validated $\lambda$ for both methods)}
         \label{fig:rpca}
\end{figure}

\begin{figure}[H]
         \centering
         \includegraphics[width=\textwidth]{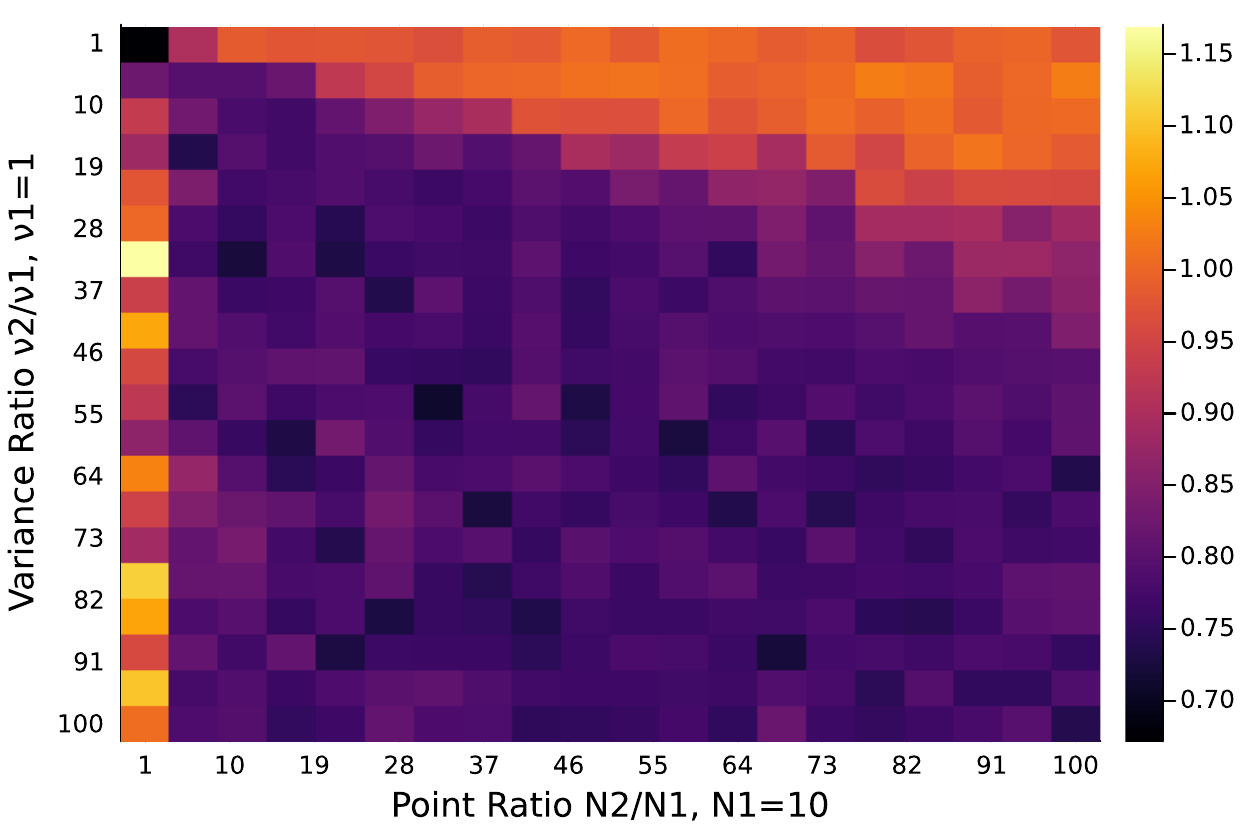}
         \caption{Ratio of subspace affinity errors \secondname/HePPCAT (unknown variance, no group knowledge, cross-validated $\lambda$ for \secondname)}
         \label{fig:heppcat}
\end{figure}

For the heatmaps in \fref{fig:pca}-\fref{fig:heppcat},
each pixel represents a ratio of \name error
divided by the error of some other method like HePPCAT.
A value close to 1 implies \name did not perform much better relative to the other method,
whereas a ratio closer to 0 implies \name performed relatively well.
For the heatmaps in \fref{fig:pca}-\fref{fig:heppcat},
the x-axis represents the point ratio between group 1 and 2,
where group 1 always has 10 points.
The y-axis represents the variance ratio between group 1 and 2,
where the group 1 noise variance was fixed to 1. The average of 25 trials is plotted at each pixel in the heatmap where each trial has different noise, basis coefficients, and subspace basis realizations.
To summarize, we explored the effects of data quality and data quantity
on the heteroscedastic subspace basis estimates 
in different situations.

\fref{fig:pca} and \fref{fig:pca_good} compare \name against PCA
in the simpler situation where noise variances are known.
\fref{fig:pca_good} is similar to \fref{fig:pca}
but only using the high quality points for PCA specifically,
whereas \name used all of the data.
This result confirms that it is useful
to use very noisy data,
as opposed to throwing it away and treating the remaining data as homoscedastic.
\fref{fig:rpca} and \fref{fig:heppcat} compare \name against two other PCA methods
in the unknown variance setting.
\fref{fig:rpca} compares against Robust PCA to see if an ``outlier'' model can capture heteroscedastic noise.
\fref{fig:heppcat} compares against HePPCAT.

\begin{figure}[H]
         \centering
         \includegraphics[width=\textwidth]{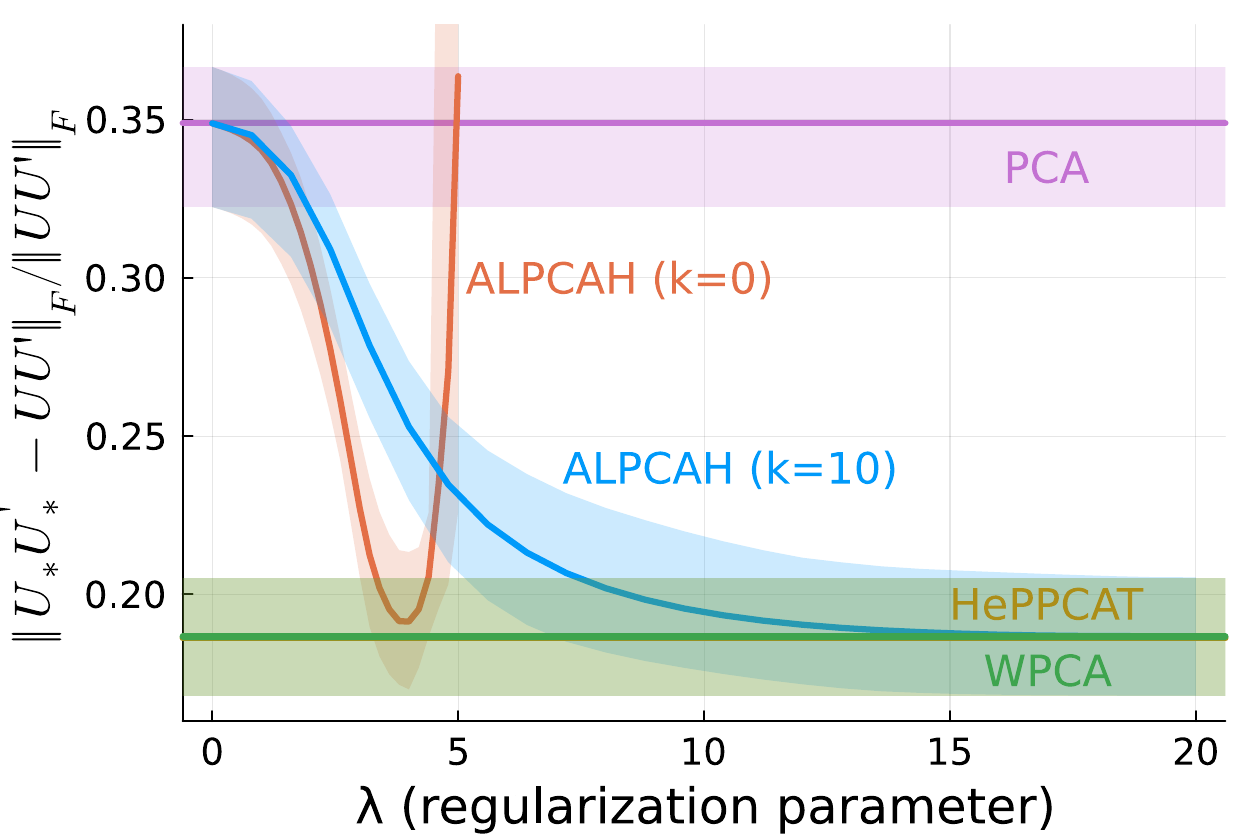}
         \caption{Subspace affinity error of various PCA methods as the regularization parameter is adjusted (known variance)}
         \label{fig:absolute_error_known}
\end{figure}

\begin{figure}[H]
         \centering
         \includegraphics[width=\textwidth]{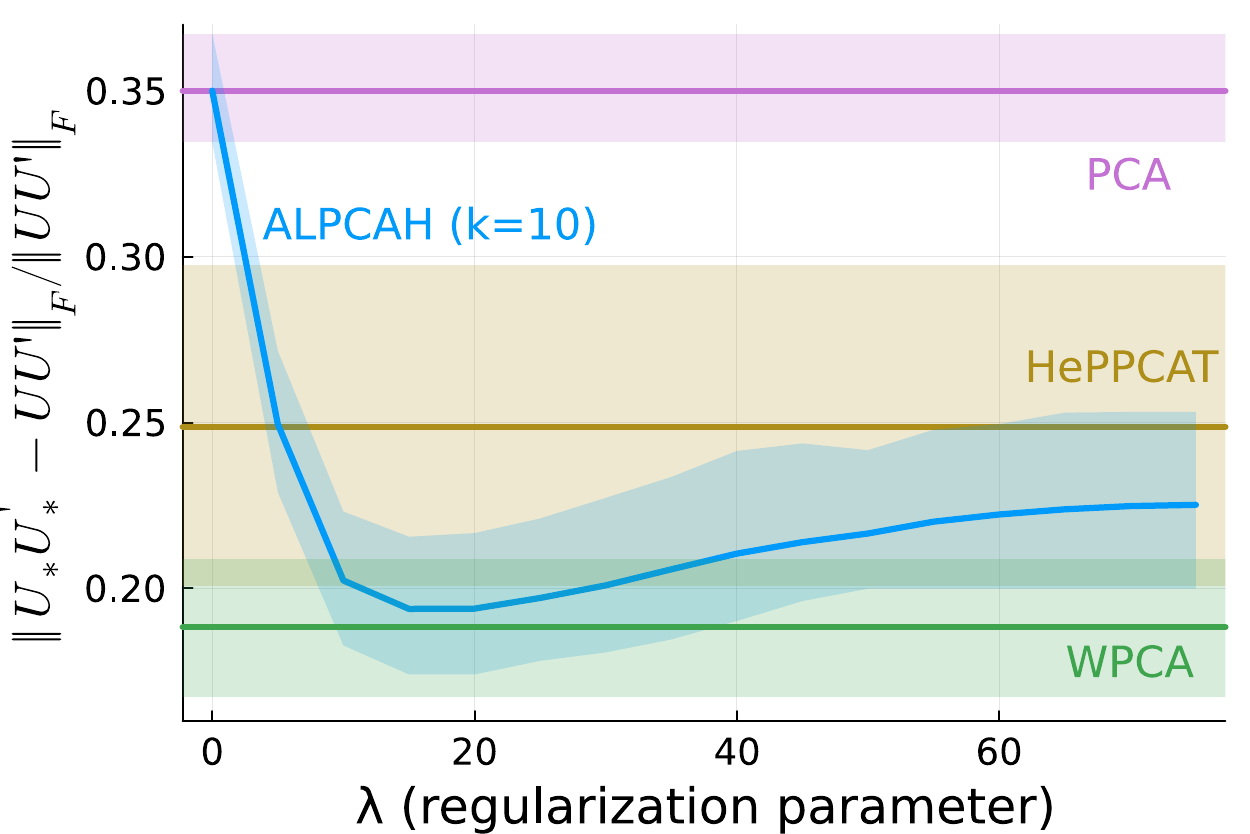}
         \caption{Subspace affinity error of various PCA methods as the regularization parameter is adjusted (unknown variance, no group knowledge)}
         \label{fig:absolute_error_unknown}
\end{figure}

Since \fref{fig:pca}-\fref{fig:heppcat} only show relative error, it is important to discuss absolute error of these algorithms. For \fref{fig:absolute_error_known}-\fref{fig:absolute_error_unknown}, we fixed the total number of points to be 500 with just enough high quality samples that have noise variance $\nu_1 = 0.25$ and the rest of the points have noise variance $\nu_2 = 100$. The regularization parameter $\lambda$ is varied both when rank knowledge is known or estimated (for these results $k=10$) and when rank knowledge is not utilized ($k=0 \implies f_k (X) = \| \cdot \|_*$). The y-axis consists of the subspace affinity error function as shown before. The average error is plotted out of 25 trials with maximum error bounds for each $\lambda$ value.
\fref{fig:absolute_error_unknown} considers the unknown variance case but shows WPCA
(a known variance method) as a bottom floor
to illustrate the lowest realistic affinity error if one knew the noise variances.

Because the unknown variance setting requires a tuning parameter $\lambda$,
we performed cross validation to determine a different $\lambda$ value for each heatmap pixel location to generate \fref{fig:rpca} and \fref{fig:heppcat}. The robustness of $\lambda$ for different point and variance ratios is mentioned in the discussion section. In practice, we found that one $\lambda$ value works well across the entire heatmap. Experimentally, we found that a sufficiently large $\lambda \geq \|Y\|_2$ gave the lowest subspace affinity error in the known variance setting so cross-validation is not performed in the known variance setting for these experiments.

Note that the subspace basis dimension was used for these results in \fref{fig:pca}-\fref{fig:heppcat} by setting $k = 10$ for $f_k (X)$ in \name to compare against HePPCAT but one may use $k=0$ when the subspace dimension is not known.
For many applications, the subspace dimension is unknown such as non-Lambertian surfaces under non-isotropic lighting conditions \cite{lambertian}.
For this situation, there are rank estimation methods proven to be robust in this heteroscedastic noise setting,
such as randomly flipping signs in the data matrix \cite{signFlips}.
Thus, it is possible to approximate the rank
beforehand given a reasonably sized data matrix such that SVD methods are feasible.

\newpage

\section{Discussion}
In the known variance cases,
\fref{fig:pca} shows that \name performs well relative to PCA in noisy situations
and can improve estimation by up to 50\% or 20\% in more tame situations.
From the bottom left corner and moving rightwards,
there is a general decline as the estimation error worsened
as the number of ``bad'' points increased.
This means that the noisy points contributed too much to the estimation process
when the good quality data should have more influence in the process.
\name uses the term
$ \normfrobr{ Z \Pi^{-1/2} }^2 $,
but a user may well use something like
$ \normfrobr{ Z \Pi^{-1} }^2 $
to further downplay the contribution of the noisy points.
Finding the optimal weighing scheme for this method is a topic of future work,
but using inverse standard deviations is a natural choice
that arises from the Gaussian likelihood.
Some work has been done in this area
for the case when noise variances are known \cite{optimal_pca}.
In \fref{fig:pca_good}, one can see that even in a limited data situation with very noisy data (bottom left corner),
there is a 30\% improvement relative to applying PCA to just the good data alone.
The improvement only increased as more noisy points were added.
Thus
it is beneficial to collect and use all of the data,
since the noisy points offer meaningful information
that can improve the estimate of the basis
versus using good data alone.

For the unknown variance cases,
we considered the situation where group information is not known
in the sense that each data point is treated as having its own noise variance
as opposed to belonging to known groups $\{1,2\}$.
This groupless situation is more challenging than the grouped case.
Because of this, it is useful for comparison purposes in this unknown variance case.
Since Robust PCA shares similarities with \secondname,
\fref{fig:rpca} compared these two methods.
As illustrated, using cross-validation to learn $\lambda$ for both methods,
\name was able to outperform RPCA in all situations shown in the heatmap.
Thus it appears to be preferable
to treat extremely noisy points
with a noise model $ \normfrob{ \cdot } $
rather than treating them as outliers
with a $ \| \cdot \|_{1,1} $ regularizer.
In
\fref{fig:heppcat},
the comparison with
HePPCAT,
there is one location (bright yellow)
where \name gave a worse estimation of the subspace basis,
but generally on average there was a 20\% improvement over HePPCAT.
Since HePPCAT is a hard rank constraint method,
it seems beneficial to not completely shrink the $k+10$ singular values
but rather to retain them
as they seem to improve the estimation process.
Moreover, since we make no distributional assumptions about $X$ itself besides low-rank assumptions,
then this assumption relaxation 
helps us achieve lower error in settings where the basis coordinates are not Gaussian,
whereas HePPCAT makes Gaussian assumptions about the basis coordinates themselves.

For both \fref{fig:absolute_error_known} and \fref{fig:absolute_error_unknown}, Robust PCA is excluded since this method did not perform any better than PCA for this specific test setup of fixed noise variances and point ratio. In \fref{fig:absolute_error_known}, for \name ($k=0$), the results became worse than PCA once $\lambda > 5$ so the y-axis range is fixed for better visibility. For this case, it is interesting that there is a certain $\lambda$ range where \name ($k=0$) performs similarly to HePPCAT. Recall that when $k=0$ in the known variance setting, the optimization problem is convex. When rank knowledge is utilized, \name ($k=10$) subspace affinity error approaches the error of the other methods as $\lambda$ grows and stays fixed at that location for larger $\lambda$ values. In \fref{fig:absolute_error_unknown}, \name ($k=0$) performs poorly in the unknown variance case and is excluded just like Robust PCA. The results shown are in the harder groupless setting where noise variance group knowledge is not known. We observed that \name ($k=10$) performs better than HePPCAT for a certain $\lambda$ range and gets close to WPCA (known variance method) as if we really did learn the noise variances of the points.

\begin{figure}[H]
         \centering
         \includegraphics[width=\textwidth]{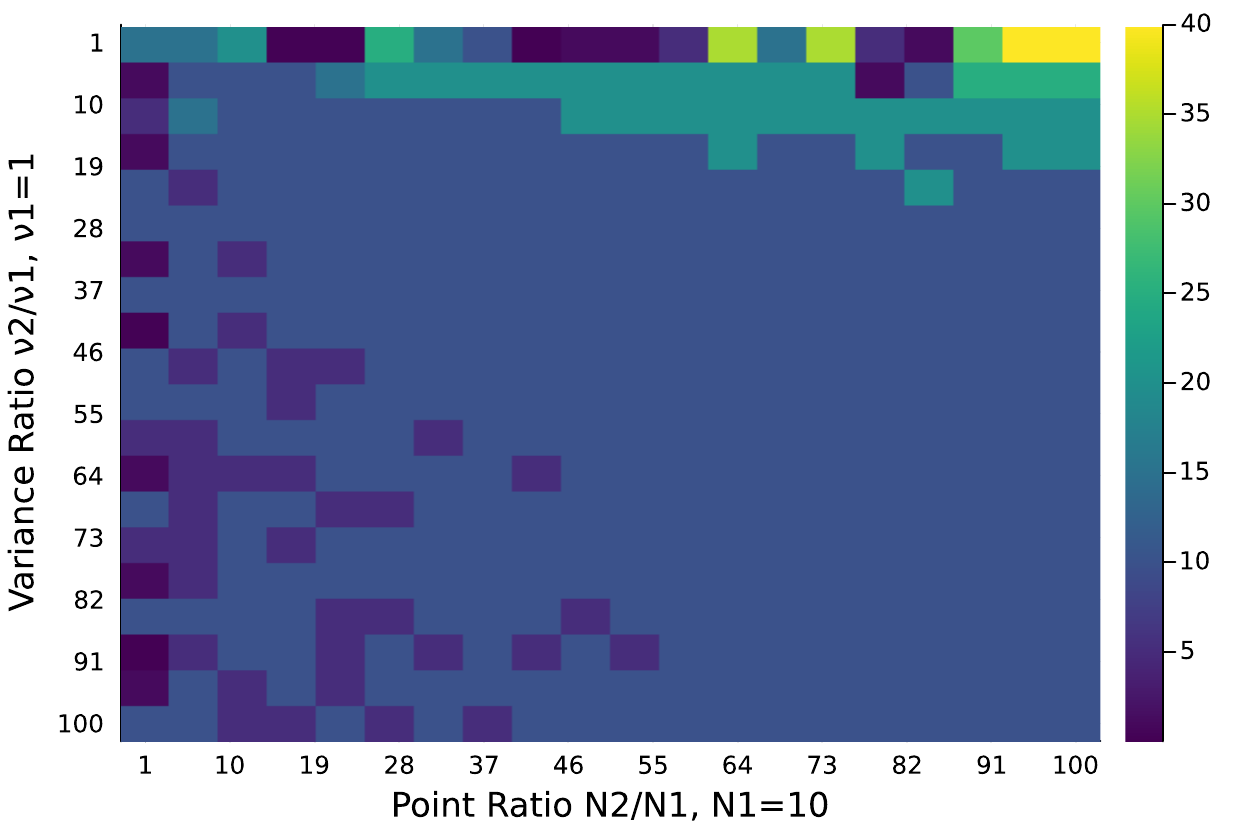}
         \caption{\name cross validation matrix of optimized $\lambda$ values}
         \label{fig:cv_matrix}
\end{figure}

As a final comment, the regularization parameter appears to be fairly robust against this landscape of different variance ratios and point ratios shown in \fref{fig:pca}-\fref{fig:heppcat}. Recall that each heatmap pixel involves a different random subspace basis with data points that have different noise and basis coordinate realizations.
The robustness of the regularization parameter
means that it will be easier to find a suitable value since,
for example, the user does not need to worry about data split size between validation and test set
or whether the variance ratio in that specific validation set will not generalize to the test set.
\fref{fig:cv_matrix}
shows this result.

\section{Conclusion}
Heteroscedastic data can
exist when using mixed data sets that stem from different sources to give an example.
Current methods to deal with subspace models in this setting
have limitations such as requiring the noise variances to be known
or assuming Gaussian basis coefficients.
Both of these assumptions may not be good assumptions in practice
due to unknown data set group knowledge or data distribution knowledge.
This work proposed a PCA method named \name
that can estimate the noise variances of the collected data
and use these estimates in the optimization model
to not only denoise the data, but also improve the estimate of the subspace basis.
\name avoids the limitations stated above
and led to higher accuracy in the subspace basis estimate
as shown in the results section. 

\newpage

\section{Funding Disclosure}
This work is supported in part by NSF CAREER Grant 1845076.
\bibliographystyle{IEEEtran}
%
\bibliography{ref.bib}


\end{document}